\documentclass[review]{elsarticle}

\usepackage{lineno,hyperref}
\usepackage[utf8]{inputenc} 
\usepackage[T1]{fontenc}    
\usepackage{url}            
\usepackage{booktabs}       
\usepackage{amsfonts}       
\usepackage{nicefrac}       
\usepackage{microtype}      
\usepackage{lipsum}
\usepackage{graphicx} 
\usepackage{verbatim}
\usepackage{color}
\usepackage{soul}
\modulolinenumbers[5]
\usepackage{cleveref}

\journal{Knowledge-Based Systems}

\renewcommand\hl[1]{#1}

\begin{document}

\begin{frontmatter}

\title{MICE: Mining Idioms with Contextual Embeddings}

\author{Tadej Škvorc}
\address{University of Ljubljana, 
  Faculty of Computer and Information Science, 
  1000 Ljubljana, Slovenia\\
  Jožef Stefan Institute,
  Jamova Cesta 39,
  1000 Ljubljana, Slovenia}
  \ead{tadej.skvorc@fri.uni-lj.si}
\author{Polona Gantar}  
\address{University of Ljubljana,
   Faculty of Arts,
  1000 Ljubljana, Slovenia }
  \ead{apolonija.gantar@guest.arnes.si}
\author{Marko Robnik-Šikonja}
\address{University of Ljubljana, 
  Faculty of Computer and Information Science, 
  1000 Ljubljana, Slovenia}
  \ead{marko.robnik@fri.uni-lj.si}




\begin{abstract}
Idiomatic expressions can be problematic for natural language processing applications as their meaning cannot be inferred from their constituting words. A lack of successful methodological approaches and sufficiently large datasets prevents the development of machine learning approaches for detecting idioms, especially for expressions that do not occur in the training set. We present an approach called MICE that uses contextual embeddings for that purpose. We present a new dataset of multi-word expressions with literal and idiomatic meanings and use it to train a classifier based on two state-of-the-art contextual word embeddings: ELMo and BERT. We show that deep neural networks using both embeddings perform much better than existing approaches and are capable of detecting idiomatic word use, even for expressions that were not present in the training set. We demonstrate the cross-lingual transfer of developed models and analyze the size of the required dataset.
\end{abstract}


\end{frontmatter}

\section{Introduction}

Idiomatic expressions (IEs), also called idioms, are composed of a group of words whose meaning is established by convention and cannot be deduced from individual words composing the expression (e.g., it's a piece of cake). 
In this work we, are interested in the detection and identification of IEs.

Due to the lack of  satisfactory tools, linguists often create lexicons of idioms manually or by using tools that take into account only co-occurrence features, since these are easier to implement and are relatively language independent. This type of workflow introduces several problems. First, manually created large lexicons of idioms  are scarce because of the time-consuming human labor that is required, particularly for less-resourced languages. Second, frequency lists of idioms that were created without robust, generalized identification tools are unreliable due to their discontinuity and syntactic variability. Finally, discovery or identification of new IEs is  often based on the personal knowledge of linguists or frequent collocations. This may completely omit many idioms.

IEs such as "break the ice" and "under the weather" commonly occur in texts. They can be hard to understand for computer models as their meaning differs from the meaning of individual words. To address this, several automatic machine learning based approaches for the detection of idiomatic language emerged. However, current approaches suffer from a number of issues and limitations related to methodological shortcomings and a lack of datasets.
The first issue that affects current approaches is the lack of large datasets with annotated IEs.
Because of a large number of different IEs, a dataset that would contain  sufficient number of examples for every IE needed to train a classification model currently does not exist. Additionally, most existing datasets only address English, which makes developing approaches for other languages difficult. Existing works use small datasets, such as the data from SemEval 2013, task 5B \cite{korkontzelos2013semeval}, PARSEME Shared Task on Automatic Verbal Multi-Word Expression (MWE) Identification \cite{savary2017parseme}, or the VNC tokens dataset \cite{cook2008vnc}. These datasets only cover a limited number of IEs and contain at most a few annotated sentences for each expression, making it hard to train successful machine-learning models for IE recognition.

Deep neural networks are currently the most successful machine learning approach for textual data, surpassing all other approaches in practically all language processing and understanding tasks \citep{lecun2015deep,Zhang2015,Kim2016,peters2018deep,devlin2018bert}. As input, neural networks require numerical data, and texts are transformed into numeric vectors via a process called text embedding. The process has to ensure that relations between words are reflected in distances and directions in a numeric space of typically several hundred dimensions.  The embedding vectors are obtained from specialized learning tasks based on neural networks, e.g., word2vec \citep{mikolov2013exploiting}, GloVe \citep{pennington2014glove}, or fastText \citep{Bojanowski2017}.  For training, the embedding algorithms use large monolingual text corpora and design a learning task that tries to predict a context of a given word.
The problem of the first generation of neural embeddings, such as word2vec, is their failure to express polysemous words. During the training of the embedding, all senses of a given word (e.g., \emph{paper} as a material, as a newspaper, as a scientific work, and as an exam) contribute relevant information about their contexts in proportion to their frequency in the training corpus. This causes the final vector to be placed somewhere in the weighted middle of all word's meanings.
Consequently, rare meanings of words (which mostly include their idioms) are poorly expressed with these embeddings and the resulting vectors do not offer good semantic representations. For example, none of the 50 closest vectors of the word \emph{paper} is related to science\footnote{A demo showing near vectors computed with word2vec from Google News corpus is available at \url{http://bionlp-www.utu.fi/wv_demo/}.}. 

The idea of contextual embeddings is to generate a different vector for each context a word appears in, and the context is typically defined sentence-wise. To a large extent, this solves the problems with word polysemy, i.e. the context of a sentence is typically enough to disambiguate different meanings of a word for humans as well as for the learning algorithms. 
 In our work, we  use what are currently the most successful approaches to contextual word embeddings, ELMo \citep{peters2018deep} and BERT \citep{devlin2018bert}. 
We examine whether contextual word embeddings can be used as a solution to the idiom identification problem. Past work shows that contextual word embeddings are capable of detecting different meanings of polysemous words and can improve the performance on a variety of NLP tasks \cite{devlin2018bert}. However, to the best of our knowledge, current approaches have not used contextual word embeddings for differentiating between idiomatic and literal language use. In the proposed approach, called MICE (Mining Idioms with Contextual Embeddings), we use ELMo and BERT embeddings as an input to a neural network and show that using them as the first layer of neural networks improves results compared to existing approaches. We evaluate our approach on a new dataset of Slovene IEs, as well as on the existing dataset from the PARSEME Shared Task on Automatic Verbal MWE Identification. To test if ELMo and BERT representations contain complementary information, we use a recent Bayesian ensemble model to combine the predictions of different MICE models. This is the first attempt to combine BERT and ELMo embeddings using an ensemble approach for idiom detection. We analyze different properties of the proposed models, such as the amount of labelled data required to get useful results, different variants of BERT models, and cross-lingual transfer of trained models.  The contributions of the paper can be stated explicitly as follows.
\begin{itemize}
    \item The first approach to use contextual embedding models (ELMo and BERT) to detect IEs.
    \item The first system to successfully recognize IEs not present in the training set.
    \item The first system to successfully analyze both sentence-level and token-level IE detection.
    \item The first successful cross-lingual approach for detection of IEs.
    \item The first Bayesian ensemble approach to combine ELMo and BERT-based models.
    \item An extensive analysis of different properties of IE detection, such as differences in the recognition rate for different IEs and different amounts of training data.
    \item Creation of SloIE, a large dataset of IEs in less-resourced, morphologically rich Slovene language.
\end{itemize}
We show that contextual embeddings contain a large amount of lexical and semantic information that can be used to detect IEs. Our MICE approach outperforms existing approaches that do not use pre-trained contextual word embeddings in the detection of IE present in the training data, as well as identification of IE missing in the training set. The latter is a major problem of existing approaches. Finally, we show that multilingual contextual word embeddings are capable of detecting IEs in multiple languages even when trained on a monolingual dataset. 

The reminder of the paper is structured as follows. In Section \ref{sec:related_work}, we describe past research on automatic IE detection. We present our MICE methodology in Section \ref{sec:methodology}. Section \ref{sec:datasets} describes the datasets used for the evaluation of our approach, which we describe in Section \ref{sec:evaluation}. Section \ref{sec:conclusion} concludes the paper.

\section{Related Work}
\label{sec:related_work}
There currently exists a variety of approaches for detecting IEs in a text, broadly divided into supervised and unsupervised methods. In supervised approaches, the problem is frequently presented as a binary classification problem where a separate classifier is trained for each idiom \cite{liu2017representations}. The disadvantage of this approach is that it scales poorly to a large number of idioms as it requires a separate training set  for each idiom.  

In recent years, several neural network approaches have been proposed. MUMULS \cite{klyueva2017neural} uses a neural network with bidirectional gated recurrent units (GRUs) \cite{cho2014learning} in combination with an embedding layer. In addition to idioms, it is capable of detecting different types of verbal multi-word expressions, which were annotated within the PARSEME Shared Task on Automatic Verbal MWE Identification \cite{savary2017parseme}. MUMULS achieved the best results on multiple languages, but the authors reported a poor classification accuracy on languages with a low amount of training data and were unable to detect expressions that did not occur in the training set. The 2018 edition of the shared task \cite{ramisch2018edition} featured several other systems based on neural networks \cite{berk2018deep, ehren2018mumpitz,borocs2018gbd} with similar outcomes to MUMULS, namely good results on several languages but low classification accuracy and $F_1$ score for languages with small training datasets and no detection of expressions that are not present in the training set. Another approach was presented by \citet{borocs2018gbd}, who use  a bidirectional long short-term memory network (biLSTM) in combination with graph-based decoding.  However, despite using neural networks, these approaches do not use pretrained contextual embeddings. Because of this, they cannot use un-annotated datasets when training their model, making it more difficult for them to make full use of contextual information in text.

The second broad group of methods for detecting idiomatic word use are unsupervised approaches.  \citet{sporleder2009unsupervised} use lexical cohesion to detect IEs without the need for a labeled dataset or language resources such as dictionaries or lexicons.  \citet{liu2018heuristically} compare the context of a word's occurrence to a pre-defined "literal usage representation" (i.e. a collection of words that often appear near literal uses of the word) to obtain a heuristic measure indicating whether a word was used literally or idiomatically. The obtained scores are passed to a probabilistic latent variable model, which predicts the usage of each word. They report average $F_1$ scores between 0.72 to 0.75 on the SemEval 2013 Task 5B \cite{korkontzelos2013semeval} and VNC tokens datasets \cite{cook2008vnc}. This is lower than the results obtained by our model on a comparable task.

A potential problem with current approaches is a lack of large annotated datasets that could be used to train classification models. \citet{liu2017representations} use the data from  SemEval 2013  Task 5B \cite{korkontzelos2013semeval}, which only contains 10 different idioms with 2371 examples. \citet{borocs2018gbd} and \citet{klyueva2017neural} trained their models on the PARSEME Shared Task on Automatic Verbal MWE Identification \cite{savary2017parseme}, which only contains a small number of idioms across 20 languages. Larger datasets exist, such as the VNC tokens dataset \cite{cook2008vnc}, which contains 2,984 instances of 53 different expressions, and the dataset presented by Fadaee et al. \cite{fadaee2018examining}, which contains 6,846 sentences with 235 different IEs in English and German. In our work, we use a larger dataset with 29,400 sentences and 75 different IEs.

Existing classification approaches require a list of idiomatic phrases with accompanying datasets on which a classifier is trained. Current approaches pay little attention to detecting idioms that do not appear in the training set, which is a much harder problem. However, due to a large number of idiomatic phrases, such use is more reflective of real-world problems. Even the unsupervised approach presented by \citet{liu2018heuristically} first manually constructs literal usage representations for each idiomatic phrase and is therefore not suitable for detecting non-listed IEs. We use contextual embeddings, which can capture semantic information without requiring labelled data for training. This allows them to detect idiomatic phrases even if they do not appear in a pre-defined list.

\section{Detecting IEs with Contextual Word Embeddings}
\label{sec:methodology}
We first describe two state-of-the-art deep neural network approaches to contextual embeddings, 
ELMo \cite{peters2018deep} and BERT \cite{devlin2018bert}, followed by the proposed neural network architectures for identification of IEs and a their Bayesian ensemble.

\subsection{ELMo contextual embeddings}
\label{sec:elmo}
ELMo (Embeddings from Language Models) \citep{peters2018deep} is a large pretrained neural language model, producing contextual embeddings and state-of-the-art results in many text processing tasks. 
The ELMo architecture consists of three layers of neurons. The output of neurons  after each layer gives one set of embeddings, altogether three sets. The first layer is the convolutional (CNN) layer operating on the character-level input. This layer is followed by two biLSTM layers
that consist of two concatenated LSTM layers. The first, left-to-right LSTM layer is trained to predict the following word based on the given past words, where each word is represented by the embeddings from the CNN layer. The second, right-to-left LSTM  predicts the preceding word based on the given following words.
Although ELMo is trained on character-level input and is able to handle out-of-vocabulary words, a vocabulary file containing the most common tokens is used for efficiency during training and embedding generation. 

In NLP tasks, a weighted average of the three embeddings is usually used. The weights for merging the representation of layers are learned during the training of the model for a specific task. Optionally, the entire ELMo model can be fine-tuned for the specific task.

In our work, we use the ELMo model that was pre-trained on a large amount of Slovene text \cite{ulcar2019high}. We take an average of the three ELMo embedding layers as the input to our prediction models. These embeddings are not fine-tuned to the specific task of idiom detection, as we wanted to evaluate how well the embeddings capture the relevant contextual information without task-specific fine-tuning. As results show, even without fine-tuning, the contextual embeddings improve performance compared to similar approaches that do not use contextual word-embeddings. Fine-tuning of the embedding layers of neural networks is left for further work. 

\subsection{BERT contextual model}
\label{sec:BERT}
BERT (Bidirectional Encoder Representations from Transformers) \citep{devlin2018bert} generalizes the idea of language models to masked language models---inspired by Cloze (i.e. gap filling)
tests---which test the understanding of a text by removing a certain portion of words that the participant is asked to fill in. The masked language model randomly masks some of the tokens from the input, and the task of the language model is to predict the missing token based on its neighbourhood. 
BERT uses transformer architecture of neural networks \citep{Vaswani2017}, which uses both left and right context in predicting the masked word and further introduces the task of predicting whether two sentences appear in a sequence. 
The input representation of BERT are sequences of tokens representing subword units. The result of pre-trained tokenization is that some common words are kept as single tokens, while others are split into subwords (e.g., common stems, prefixes, suffixes---if needed down to a single letter token). The original BERT project offers pre-trained English, Chinese, and multilingual models; the latter, called mBERT, is trained on 104 languages simultaneously. BERT has shown excellent performance on 11 NLP tasks:  8 from GLUE language understanding benchmark \citep{Wang2018}, question answering, named entity recognition, and common-sense inference.

Rather than training an individual classifier for every classification task from scratch, which would be resource and time expensive, the pre-trained  BERT language model is usually used and fine-tuned on a specific task. This approach is common in modern NLP because large pretrained language models extract highly relevant textual features without task-specific development and training. Frequently, this approach also requires less task-specific data. 
During pre-training, the BERT model learns relations between sentences (entailment) and between tokens within a sentence. This knowledge is used during training on a specific downstream task \citep{devlin2018bert}.
The use of  BERT for a token classification task requires adding connections between its last hidden layer and new neurons corresponding to the number of classes in the intended task. To classify a sequence, we use a special [CLS] token that represents the final hidden state of the input sequence (i.e. the sentence). The predicted class label of the [CLS] token corresponds to the class label of the entire sequence.  The fine-tuning process is applied to the whole network, and all of the parameters of BERT and new class-specific weights are fine-tuned jointly to maximize the log-probability of the correct labels. 

In our use of BERT models, we did not fine-tune the embedding weights but left them as they were after the original pre-training. This simplification significantly reduces the computational load but leads to a potential loss of accuracy. This is a possible improvement to be tested in future work, as fine-tuning the embeddings would likely improve the results.

\subsection{The proposed MICE architecture}
Our approach is based on contextual word embeddings, which were designed to deal with the fact that a word can have multiple meanings. Instead of assigning the same vector to every occurrence of a word, contextual embeddings assign a different vector to each word occurrence based on its context. As the contexts of words' literal use and  idiomatic occurrences of the same word are likely to differ, these embeddings shall be well-suited for detecting IEs. We used two state-of-the-art embedding approaches: ELMo \cite{peters2018deep} and BERT \cite{devlin2018bert}. For ELMo, we used the pretrained Slovene model described by \cite{ulcar2019high}. The model was trained on the Gigafida corpus \cite{krek2016nadgradnja} of Slovene texts. For BERT embeddings, we use two different models: 
\begin{enumerate}
    \item The multilingual mBERT model presented by Devlin et al. \cite{devlin2018bert}, which was trained on Wikipedia text from 104 languages, including Slovene.
    \item The trilingual CroSloEngual BERT presented by \citet{ulcar2020xlbert}, which was trained on English, Slovene, and Croatian using Wikipedia for English text, the Gigafida corpus for Slovene text, and a combination of hrWaC \cite{ljubevsic2011hrwac}, articles from the Styria media group, and Riznica corpora \cite{cavar2012riznica} for Croatian text.  This BERT is better suited for classification tasks in Slovene and Croatian as mBERT as its training incorporated larger amounts of training data and a larger vocabulary for each of the involved languages. The authors also report improved cross-lingual transfer of trained models between the three languages.
\end{enumerate}

We use the embeddings (ELMo or BERT) as the first layer of a neural network. This layer is followed by a bidirectional gated recurrent unit (GRU) with 100 cells. GRUs are similar to standard recurrent units but use an additional update and reset gate to help deal with the vanishing gradient problem. The update gate is defined as
\begin{equation}
    z_t = \sigma (W^{(z)}x_t + U^{(z)}h_{t-1} + b_z),
\end{equation}
where $W^(z)$ and $U^(z)$ are trainable weights, $x_t$ is the input vector and $b_z$ is the trainable bias. $h_{t-1}$ represents the memory of past inputs computed by the network. The reset gate uses the same equation, with different weights and biases:
\begin{equation}
    r_t = \sigma (W^{(r)}x_t + U^{(r)}h_{t-1} + b_r).
\end{equation}
For each input, the GRU computes the output as:
\begin{equation}
    h_t = z_t \odot h_{t-1} + (1-z) \odot \tanh (W^{(h)} x_t + U^{(h)} (r_t \odot h_{t-1}) + b_h),
\end{equation}
where $\odot$ is the Hadamard product, and $W^{(h)}$, $U^{(h)}$, and $b_h$ are trainable weights and biases.

For both ELMo and BERT embeddings, we follow the GRU layer with a softmax layer to obtain the final predictions. A dropout of 50\% is applied at the softmax layer. This approach follows the work on MWE detection presented by \citet{klyueva2017neural} but with the difference that we use contextual embeddings. We deliberately use a simple network architecture to show that the embeddings, by themselves, capture enough semantic information to properly recognize IEs.  

We use the architecture on two types of classification tasks: a token-level classification, where we predict whether an individual token has an idiomatic or literal meaning, and a sentence-level classification, where the network makes a single prediction for the entire sentence, predicting whether the sentence contains an expression with an idiomatic meaning. The details of the tasks are presented in \Cref{sec:evaluation}.

We fine-tuned the hyperparameters using a development set consisting of 7\% of sentences randomly selected from our dataset, as described in  \Cref{sec:monolingual-dataset}. We trained the network for 10 epochs using RMSProp as the optimizer with the learning rate of 0.001, $\rho=0.9$, and $\epsilon=10^{-7}$. We used the binary cross-entropy as the loss function.

\subsection{Bayesian ensemble of MICE models}
\label{sec:BayesianEnsemble}
\hl{Due to the fact that MICE can be used with different embeddings, it is possible that the information extracted by different embedding models is complementary. In this case, an ensemble of different embedding models could improve the performance by learning their combination that best suits each idiom. We test this hypothesis using the Multivariate Normal Mixture Conditional Likelihood Model (MM), a Bayesian ensemble model proposed by} \citet{pirs2019bayesian}. \citet{miok2020bayesian} \hl{showed that an MM ensemble can improve the performance of individual classifiers on the text annotation task. We test this approach on the IE detection task and combine our three best models (MICE with Slovene ELMo, MICE with mBERT, and MICE with CroSloEngual BERT)}. \hl{We transform the predictions of each model using an inverse logistic transformation and concatenate them to obtain a $(m-1)r$-variate distribution, where $m=2$ is the number of classes and $r=3$ is the number of models. As explained below, we model the latent distribution using the multivariate normal mixtures conditional on the labels and predictions obtained from the training set in a similar fashion to linear discriminant analysis. We then generate predictions on the test data using the following formula:}

$$
p(T^* = t|u^*, \theta) =  \frac{p(u^*|\theta_t)(\gamma_t n_t)}{\sum^r_{i=1}p(u^*|\theta_i) (\gamma_i n_i)},
$$
\hl{where $p$ is the probability density function, $\gamma_t$ is the frequency prior for class $t$, $n_t$ is the number of true labels in class $t$ from the training dataset, $\theta$ are the estimated parameters of the Bayesian model, $\theta_t$ are the subset of the parameters with the true label $t$, $T^* \in \{1, 2, ..., m\}$ is the response random variable corresponding to the predicted class, and $u^*$ is the concatenated and transformed probability vectors of our embedding models. The model is described in detail in} \citet{miok2020bayesian}. \hl{As a baseline ensemble, we use simple (unweighted) voting.}

\hl{We evaluate the ensemble models in the same manner as the individual models. The results and discussion of this evaluation are presented in Section} \ref{sec:evaluation}.

\section{Datasets}
\label{sec:datasets}
Our approach supports two types of tasks, monolingual and multilingual. The monolingual approach requires a reasonably large dataset with a sufficient number of idioms. We analyze the required size of a dataset  in terms of different idioms and examples of their usage in  both monolingual and multilingual settings in \Cref{sec:test_dataset_size}. The multilingual approach exploits the existing monolingual dataset to transfer the trained model to languages with fewer resources, i.e. with non-existent or smaller datasets.

In Section \ref{sec:monolingual-dataset} we  describe our monolingual Slovene dataset. In Section \ref{sec:parseme} we describe the well-known PARSEME datasets \cite{savary2017parseme} for detection of multi-word expressions in many languages, which also include idioms.  

\subsection{Monolingual dataset}
\label{sec:monolingual-dataset}
We evaluate our approach on a new dataset of Slovene IEs, called SloIE, which we make publicly available for further research\footnote{http://hdl.handle.net/11356/1335}. The dataset consists of 29,400 sentences extracted from the Gigafida corpus \cite{krek2016nadgradnja} and contains 75 different IEs. The 75 IEs were selected from the Slovene Lexical Database \cite{gantar2011slovene} and had to meet the condition that they appear in corpus sentences in both idiomatic and literal senses, such as, e.g., break the ice, step on someone's toes. 
Manual selection of idiomatic examples showed that about two-thirds of the idioms in the Slovene Lexical Database (2,041 in total) that occur in both idiomatic and literal senses occur in 50\% or more of the corpus sentences in their idiomatic sense, and about one-third of the idioms occur in 50\% or more of the corpus in their literal sense, either because literal use is not possible, or it's very rare, although possible in terms of syntax and semantics (e.g. get under someone's skin). Although this finding is interesting from a (socio)linguistic point of view, in designing the dataset for our purposes, we limited ourselves to idioms that meet the condition of appearing in both the idiomatic and the literal sense in the corpus sentences, assuming that speakers can identify both the literal and the idiomatic interpretation of a term based on the context. 

Two annotators, students of linguistics, marked the complete set of 29,400 sentences. They had four possible choices: YES (the expression in a particular sentence is used in the idiomatic sense, NO (the expression is used in the literal sense), DON'T KNOW (not sure whether the expression is used in a literal or idiomatic sense) and VAGUE, (literal or idiomatic use cannot be inferred from the sentence). Student annotators were previously briefed with short instructions and provided with a sample of good examples. 
For the training of classification models, we selected only sentences where both annotators agreed on the annotation. We also disregarded examples that were marked as "vague" or "don't know". The inter-annotator agreement across the entire dataset was 0.952. 

\begin{figure}[p]
  \centering
  \hspace*{-0.15\textwidth}\includegraphics[width=1.3\textwidth]{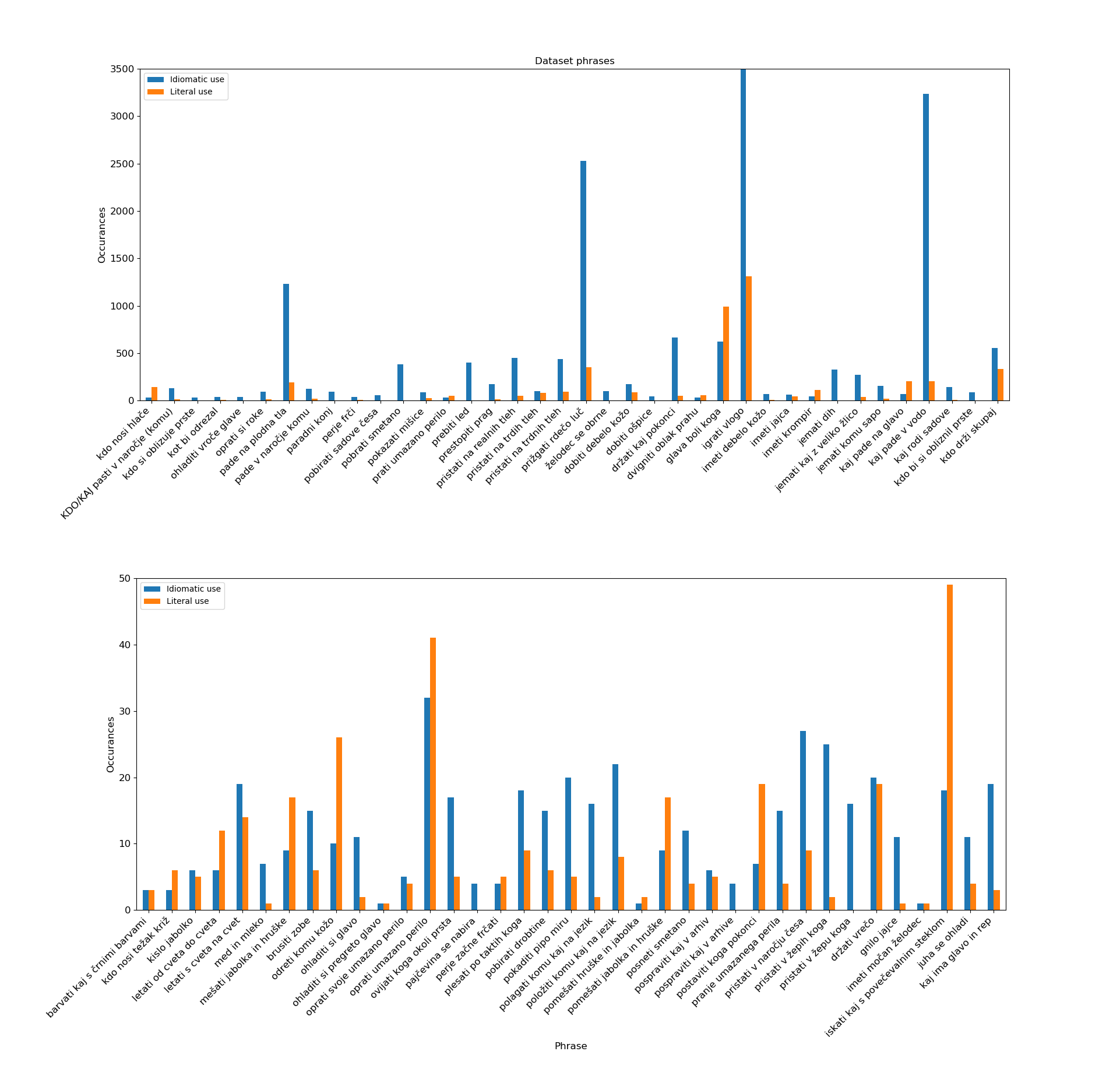}
  \caption{The number of literal and idiomatic uses for IEs present in the SloIE dataset. The top figure shows IEs that occur more than 35 times with an idiomatic meaning. The bottom figure shows IEs that occur less than 35 times with an idiomatic meaning.}
  \label{fig:dataset_both}
\end{figure}

Due to the nature of IEs, our dataset is imbalanced. A few expressions occur proportionally in both literal and idiomatic use, while most expressions occur predominately idiomatically. The dataset contains fewer than 100 occurrences for most expressions. Table \ref{tab:dataset_overview} shows an overview of the data present in our dataset. The distribution of literal and idiomatic uses of each expression is shown in Figure \ref{fig:dataset_both}.

\begin{table} [ht]
 \caption{An overview of the data present in the SloIE dataset.}
  \centering
  \begin{tabular}{lr}
    \toprule
    Sentences              & 29,400 \\
    Tokens                 & 695,636 \\
    Idiomatic sentences   & 24,349 \\
    Literal sentences      & 5,051 \\
    Idiomatic tokens & 67,088\\
    Literal tokens & 626,707\\
    Different IEs  & 75 \\
    \bottomrule
  \end{tabular}
  \label{tab:dataset_overview}
\end{table}


Table \ref{tab:dataset_ten_examples} shows ten randomly-chosen idioms from the SloIE dataset. Seven of the chosen idioms appear more often idiomatically, while three appear more often literally.

\begin{table} [h!tb]
\caption{An example of 10 idioms from the dataset with their direct/idiomatic English translations. The columns show the percentages of idiomatic ("yes"), literal ("no"), and ambiguous ("dn" $\approx$ "don't know") sentences for each idiom.}
\centering
\begin{tabular}{lcccccc}
\toprule
&\multicolumn{3}{c}{Annotator 1}& \multicolumn{3}{c}{Annotator 2}\\
Expression & yes & no & dn& yes & no & dn\\
\midrule
barvati kaj s črnimi barvami&	50&	50&	0&	50&	50&	0\\
\multicolumn{7}{l}{\textit{paint with dark colors / present pessimistically }} \\
kdo nosi hlače&	19&	75&	5&	10&	70&	19\\
\multicolumn{7}{l}{\textit{to wear pants / to be in charge }} \\
kdo nosi težak križ	&41&	50&	8&	41&	50&	8\\
\multicolumn{7}{l}{\textit{to carry a heavy cross  / to be under pressure }} \\
kdo pade v naročje&	77&	13&	9&	63&	13&	22\\
\multicolumn{7}{l}{\textit{to fall into someone's lap / to achieve with ease }} \\
kdo si oblizuje prste&	84&	4&	11&	51&	17&	31\\
\multicolumn{7}{l}{\textit{to lick ones fingers / to be very satisfied }} \\
kislo jabolko&	37&	31&	31&	25&	56&	18\\
\multicolumn{7}{l}{\textit{sour apple / unpleasant matter }} \\
kot bi odrezal&	59&	31&	9&	45&	3&	51\\
\multicolumn{7}{l}{\textit{as if cut off / instantly }} \\
letati od cveta do cveta&	30&	60&	10&	30&	40&	25\\
\multicolumn{7}{l}{\textit{fly from flower to flower / select without a plan }} \\
med in mleko&	46&	6&	46&	40&	6&	53\\
\multicolumn{7}{l}{\textit{honey and milk / abundance }} \\
oprati si roke&	85&	10&	3&	66&	14&	19\\
\multicolumn{7}{l}{\textit{wash ones hands / to be innocent }} \\
\bottomrule
\end{tabular}
\label{tab:dataset_ten_examples}
\end{table}



SloIE is much larger than other existing datasets in terms of the number of sentences. For comparison, the closest to SloIE is the VNC tokens dataset that contains 2984 instances of 53 IEs. We do not expect that datasets of such sizes would appear soon for most other languages. For that reason, we analyze the size and distribution required for successful IE detection models in Section \ref{sec:test_dataset_size}. The results could be useful guidelines for creators of similar datasets in other languages.

\subsection{PARSEME datasets}
\label{sec:parseme}
The dataset for the Edition 1.1 of the PARSEME shared task on automatic identification of verbal multiword expressions (MWEs) consists of  280,838 annotated sentences split across 20 languages. The corpus contains annotations for various types of verbal MWEs, such as verb-particle constructions, inherently reflexive verbs, and verbal idioms. As our work focuses on detecting IEs, we only predict tags of verbal idioms. A summary of the number of sentences for each language used in our work is presented in Table \ref{tab:parseme_overview}. We do not use the Arabic dataset as it was not made available under an open licence.

\begin{table} [h!t]
 \caption{An overview of the data present in the PARSEME datasets. Of the 20 languages in the PARSEME corpus, we use 18. We omit Arabic because it is not available as an open language and Farsi, which does not contain IEs. On average, each language contains 586 IEs.}
  \centering
  \begin{tabular}{lccc}
    Language & Sentences & Tokens & IEs \\
    \toprule
    BG & 6,913  & 157,647 & 417   \\
    DE & 6,261  & 120,840 & 1,005 \\
    EL & 5,244  & 142,322 & 515   \\
    EN & 7,436  & 124,203 & 59    \\
    ES & 2,502  & 102,090 & 196   \\
    FA & 2,736  & 46,530  & 0     \\
    FR & 17,880 & 450,221 & 1,786 \\
    HE & 4,673  & 99,790  & 86    \\
    HU & 3,569  & 87,777  & 92    \\
    HR & 3,003   & 69915   & 131   \\
    IT & 15,728 & 387,325 & 913   \\
    LT & 12,153 & 209,636 & 229   \\
    MT & 5,965  & 141,096 & 261   \\
    PL & 11,578 & 191,239 & 317   \\
    PT & 19,640 & 359,345 & 820   \\
    RO & 45,469 & 778,674 & 524   \\
    SL & 8,881  & 183,285 & 283   \\
    SV & 200    & 3,376   & 9     \\
    TR & 16,715 & 334,880 & 2,911 \\ \hline
    Total & 19,6546 & 3,990,191 & 10,554 \\
    \bottomrule
  \end{tabular}
  \label{tab:parseme_overview}
\end{table}

IEs in the PARSEME datasets only occur in a small number of sentences. Additionally, most IEs occur only once in the corpus, which makes training a classifier difficult.
For that reason, we used the PARSEME dataset to evaluate our cross-lingual model. The model used the pretrained mBERT embeddings from \cite{devlin2018bert},  was further trained on our Slovene SloIE dataset, and tested on each of the PARSEME datasets in different languages. The details are reported in \Cref{sec:XLevaluation}.

\section{Evaluation}
\label{sec:evaluation}
We evaluate our MICE approach in five different settings, explained below. We present the results of these evaluation scenarios in the subsections.  

\begin{enumerate}
\item \emph{Classification of IEs that were present in the training set.} In \Cref{sec:inTraining}, we evaluate whether MICE is capable of detecting IEs that were present in the training set. This task is easier than detection of IEs not present in the training set, but still difficult due to the fact that idioms in the SloIE dataset can appear both literally or idiomatically. An English example would be the phrase "breaking the ice", which can have both the literal meaning ("The ship was breaking the ice on its way across the Arctic) or the idiomatic meaning ("I had trouble breaking the ice at the party). The models for this task have to recognize the meaning of the phrase based on its context. We split this task into two sub-tasks: i) sentence-level classification, where the network makes a single prediction for the entire sentence, predicting whether that sentence contains an expression with an idiomatic meaning  and ii) token-level classification, where we predict whether each token has a literal or idiomatic meaning. The sentence-level classification task is easier, but the token-level task can be more useful, as it can be used to detect which tokens have the idiomatic meaning.

\item \emph{Classification of IEs that were not present in the training set.} Due to a large number of idioms, it is difficult and expensive to annotate a dataset that would cover every idiom. Because of this, it would be desirable that the prediction model is capable of detecting expressions that are not present in the training set. We test this setting in \Cref{sec:outsideTraining}. As with the first task, we use sentence-level and token-level classification. This task is more difficult than detection of IEs present in the dataset, and can only be solved successfully if the contextual word embeddings contain information about idiomatic word use (e.g., as directions in the vector space). 

\item \emph{Difference in detection of individual IEs.}
It is possible that success in detection of different IEs differs significantly, where some IEs are easy and other much more difficult to detect. In \Cref{sec:individualIEs} we evaluate how well our model detects each IE in our dataset and present the differences. 

\item \emph{Cross-lingual transfer on the PARSEME dataset.} In \Cref{sec:XLevaluation} we evaluate whether our approach can be used to detect expressions in different languages when trained with multilingual word embedding models. For testing this hypothesis, we use 18 languages from the multilingual PARSEME dataset.

\item \emph{Required size of a dataset.} Our dataset is significantly larger than other datasets used for automatic idiom detection, e.g., PARSEME (for a single language). In \Cref{sec:test_dataset_size} we conduct a series of experiments that provide an information how large dataset (in terms of number of IE and number of examples per IE) is actually needed for successful detection of idioms. This information may be valuable for other languages where similar detection tools will be built.
\end{enumerate}

We compare the proposed MICE approach to different existing approaches. As a baseline, we use the SVM classifier with the tf-idf weighted vector of a sentence as the input. We compare our approach to MUMULS \cite{klyueva2017neural}, which uses a similar neural network architecture to our approach but does not use pretrained contextual word embeddings. Unlike our approach, MUMULS uses part-of-speech tags and word lemmas as additional inputs.

\hl{For the token-based evaluation, ELMo and BERT models use different tokenization strategies: ELMo uses words as tokens, while BERT splits words into sub-word units. In the case of BERT, we use the prediction of the first token as the prediction for the entire work, following the methodology presented by} \citet{devlin2018bert}. \hl{This ensures that the token-level results are comparable between models using different tokenization strategies. }

\hl{To test our Bayesian ensemble model MICE-MM, we first train each of the ensemble members (i.e. individual MICE models) the same way as in the individual evaluation. We combine the models' predictions using the MM model. As a baseline ensemble model, we use voting (MICE-voting).}

For all tests, we report the classification accuracy (CA) and $F_1$ score. As many of the tasks are highly imbalanced, CA is not a good measure and  we mostly use the obtained $F_1$ scores in interpretations of results.

\subsection{IEs from the training set}
\label{sec:inTraining}
For the first experiment, detection of IEs present in the training set, we randomly split the SloIE dataset into training, testing, and development sets with the ratio of 63:30:7 (18,522, 8,820, and 2,058 sentences). The network was trained for 10 epochs using RMSProp as the optimizer with a learning rate of 0.001, $\rho=0.9$, and $\epsilon=10^{-7}$. Binary cross-entropy was used as the loss function. The evaluation on the development set showed that training the model for more than 10 epochs led to overfitting, likely due to the size of the dataset. We report two sets of results: recognition of individual tokens in a sentence as idiomatic or non-idiomatic (i.e. token-level classification), and detection of the whole sentence as either containing or not containing idioms (i.e. sentence-level classification).

The results for token-level classification are presented in Table \ref{tab:results_base_tokens}. To provide a sensible context for token-based classification, the input of the SVM classifier consists of the target token and three words before and three words after the target word. The SVM classifier obtains better $F_1$ score than MUMULS but lower score compared to MICE variants.  The dataset is highly imbalanced, with 96,7\% of all tokens being non-idiomatic. Lacking discriminating information, MUMULS predicts almost every token as non-idiomatic, which results in high classification accuracy but a very low $F_1$ score. Due to the imbalanced nature of the dataset, the $F_1$ score is more reflective of relevant real-world performance, and here the MICE variants are in the class of their own.

\begin{table}[htb]
 \caption{Comparison of results when classifying tokens with the same IEs present in the training and testing set. Each token was classified as either belonging to IE with the literal meaning, belonging to IE with the idiomatic meaning, or not belonging to IE.}
  \centering
  \begin{tabular}{lrr}
    \toprule
    Method     & CA  & $F_1$ \\
    \midrule
    Default classifier & 0.903 & 0.176\\
    SVM baseline & 0.8756 & 0.3962 \\
    MUMULS       & 0.975 & 0.0659  \\ 
    MICE with Slovene ELMo  & \textbf{0.981} & \textbf{0.912}  \\
    MICE with mBERT & 0.974 & 0.869  \\ 
    MICE with CroSloEngual BERT & 0.972 & 0.872 \\
    MICE-voting & 0.979 & 0.904\\
    MICE-MM &  0.979 & 0.907\\
    \bottomrule
  \end{tabular}
  \label{tab:results_base_tokens}
\end{table}


Of the MICE approaches, the one with the Slovene ELMo model obtains the highest $F_1$ score. The MICE variants with BERT embeddings obtain lower classification accuracies and $F_1$ scores. \hl{This is likely due to the fact that our ELMo embeddings were pretrained on a large amount of only Slovene texts, while the mBERT  model was trained on 104 different languages.}  Only a small amount of Slovene texts was included in its training and it has a small proportion of Slovene words in the vocabulary. The CroSloEngual embeddings were trained on a larger amount of Slovene text and therefore achieve better results. 

\hl{In token-level classification, the MICE-MM ensemble does not outperform the best individual model (Slovene ELMo embeddings). However, a separate MM ensemble model, trained only on the two BERT models, outperforms each individual BERT model (CA=0.976, $F_1=0.888$). A possible explanation of this result is that the Slovene ELMo model outperforms the two BERT models to such an extent that they do not contribute any additional information to the ensemble.}

In the evaluation on the sentence-level, instead of classifying each token, we classified each sentence \hl{based} on  whether it contains a IE or not. This lowers the importance of different tokenization strategies between ELMo and BERT. However, sentence-level evaluation does not show whether the approaches are capable of detecting specific words in a sentence as idioms. The results of this evaluation are presented in Table \ref{tab:results_base_sentences}.

\begin{table}[htb]
 \caption{Comparison of results when classifying sentences from the SloIE dataset and the same IEs are present in the training and testing sets. Each sentence was classified as either containing an expression with the literal meaning or containing an expression with the idiomatic meaning.}
  \centering
  \begin{tabular}{lcc}
    \toprule
    Method     & CA  & $F_1$\\
    \midrule
    Default classifier & 0.828 & 0.906 \\
    SVM baseline & 0.900  & 0.942 \\
    MUMULS       & 0.915  & 0.948 \\
    MICE with Slovene ELMo   & 0.951  & 0.980\\
    MICE with mBERT   & 0.897  & 0.908 \\
    MICE with CroSloEngual BERT & 0.921 & 0.954\\
    MICE-voting & 0.964 &  0.979\\
    MICE-MM & \textbf{0.971} & \textbf{0.982} \\

    \bottomrule
  \end{tabular}
  \label{tab:results_base_sentences}
\end{table}

The sentence-level classification task is less difficult, which leads to an improved performance for all models. The SVM baseline outperforms the mBERT model. MUMULS also achieves better results, outperforming the SVM baseline and the mBERT approach. MICE with CroSloEngual BERT is closer to ELMo in this task, though the latter still achieves the best scores. MICE with mBERT likely achieves lower scores because this model was not pretrained on a large enough amount of Slovene text.
\hl{Unlike for token-level classification, the sentence-level results show that the Bayesian ensemble improves the performance. This indicates that MICE models with Slovene ELMo and with the two BERT models contain complementary information which can be generalized from the train to test set. The Bayesian ensemble model learned which combination of the individual models is best suited for each idiom and increased the classification accuracy and $F_1$ score compared to individual models and the voting baseline.}

\hl{The results confirm the assumption that for different IEs, different embedding models perform best. A further analysis shall determine why this occurs in sentence-level and not in  token-level classification. In future work, we plan to conduct a comprehensive quantitative and qualitative analysis using a larger number of embedding models to determine the impact of embeddings on different idioms.}

\subsection{IEs outside the training set}
\label{sec:outsideTraining}
In the previous experiment with the same IEs present in both the training and testing set, we were able to obtain good results (especially with our contextual embeddings approach). 
However, many languages lack large annotated datasets and even when they do exist, they are unlikely to contain every possible IE found in that language. Because of this, evaluations containing IEs in both sets over-estimates the practical importance of tested methods. 

To address this, we tested how well the approaches based on contextual word embeddings generalize to IEs outside the training set. For this experiment, we split our dataset into a training and testing set so that IEs from the testing set do not appear in the training set. Apart from this change, everything else remained the same as in \cref{sec:inTraining} above.  

Since IEs in the test set are not present in the training set, the classification models cannot learn how to detect them based on word-data alone. We hypothesize that their detection is possible based on contexts in which they appear. As the meaning of an IE is different from the literal meaning of its constituting words, it should appear in a different context. Neural networks with contextual word embeddings could detect such occurrences.
Indeed, our results for token- and sentence-level IE detection, presented in Tables \ref{tab:results_hard_tokens} and \ref{tab:results_hard_sentences}, show that approaches that do not use contextual word embeddings fail to successfully detect IEs that did not occur in the training set, while MICE  approaches using contextual embeddings extract useful information.

For token level results, shown in Table \ref{tab:results_hard_tokens}, due to the imbalanced class distribution, all approaches \hl{except for MICE-Voting and MICE-MM} lag behind the default classifier concerning CA. 
For both the SVM baseline and MUMULS this is the case also in terms of $F_1$ score. The MICE approach with ELMo and mBERT models manages to correctly classify a number of IEs, though the results are worse than in the  scenario where the same IEs are present in both the training and testing set. MICE with ELMO embeddings is again the best method, while CroSloEngual embeddings are surprisingly unsuccessful.

\begin{table}[htb]
 \caption{Comparison of results when classifying tokens and test set IEs are not present in the training set. }
  \centering
  \begin{tabular}{lrr}
    \toprule
    Method     & CA & $F_1$ score\\
    \midrule
    Default classifier & 0.903 & 0.176\\
    SVM baseline & 0.870 & 0.029    \\
    MUMULS       & 0.873 & 0.000      \\
    MICE with Slovene ELMo & 0.803 & \textbf{0.866}  \\ 
    MICE with mBERT   & 0.733 & 0.803  \\ 
    MICE with CroSloEngual BERT & 0.759 & 0.176 \\ 
     MICE-voting & 0.917 &  0.599\\
     MICE-MM & \textbf{0.925} & 0.662 \\
    \bottomrule
  \end{tabular}
  \label{tab:results_hard_tokens}
\end{table}

Sentence-level results in Table \ref{tab:results_hard_sentences} show improved scores of all models. The SVM baseline and MUMULS still lag behind the default classifier concerning both CA and $F_1$ score. 
MICE approaches are better, with Slovene ELMo variant achieving the best scores.

\begin{table}[htb]
 \caption{Comparison of results when classifying sentences and the test set IEs are not present in the training set. }
  \centering
  \begin{tabular}{lcc}
    \toprule
    Method     & CA & $F_1$ score\\
    \midrule
    Default classifier & 0.828 & 0.906 \\ 
    SVM baseline & 0.783 & 0.689 \\
    MUMULS       & 0.520 & 0.672 \\
    MICE with Slovene ELMo & \textbf{0.936} & \textbf{0.964} \\
    MICE with mBERT   & 0.888 & 0.939 \\
    MICE with CroSloEngual BERT & 0.914 &  0.952\\
    MICE-voting &  0.915 & 0.953\\
    MICE-MM &    0.934 & 0.963 \\
    \bottomrule
  \end{tabular}
  \label{tab:results_hard_sentences}
\end{table}

\hl{When evaluating our approach on models outside the training set, the MICE-MM ensemble model is unable to improve the performance in terms of the $F_1$ score. This can be explained by the fact that the MM approach uses the predictions made on the training data IEs to learn the latent distributions of IE predictions from the test data, which might not generalize well. The voting ensemble MICE-voting also does not improve performance over the best-performing individual model.}

\subsection{Evaluation of individual IEs}
\label{sec:individualIEs}
\begin{figure}[htb]
  \centering
  \includegraphics[width=0.99\textwidth]{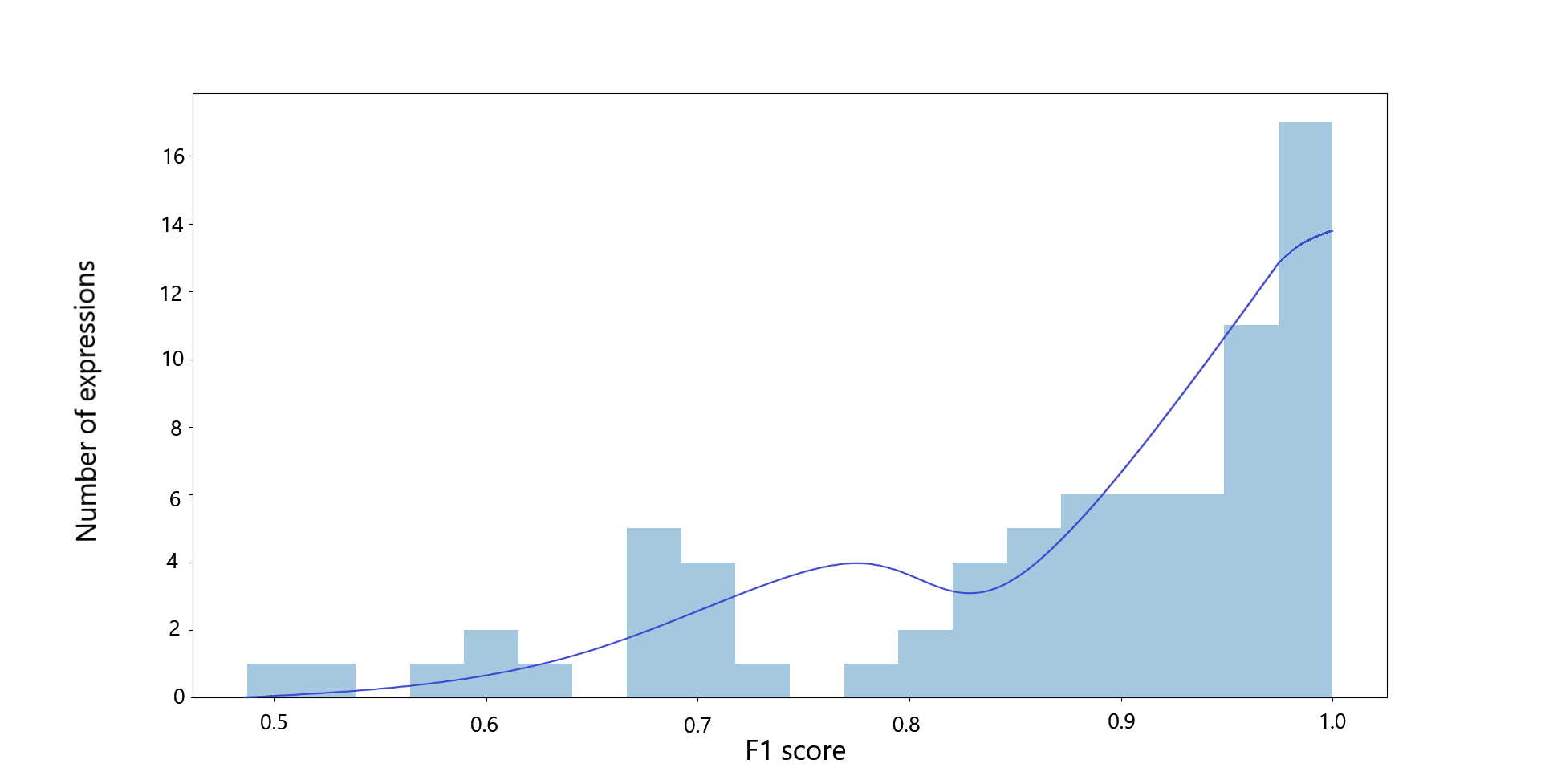}
  \caption{The distribution of $F_1$ scores per IEs in sentence-level task on the out-of-test-set task using MICE with Slovene ELMo embeddings. 
  }
  \label{fig:f1_scores}
\end{figure}

In addition to cumulative results of the entire test set, we are also interested in individual differences between IEs, as it is possible that some IEs are easy and others are hard to detect. As the meanings of IEs can vary from being similar or very different to the literal meanings of their words, we assume that the ability of models based on contextual word embeddings could vary significantly.  For this task, we train the detection models on all other IEs (74 of them) and test them on the left-out IE. In this way, we obtain a separate detection model for each IE, trained on every sentence that did not contain that IE, and evaluate it on the sentences containing that IE (similar out-of-test-set sentence-level scenario as in \Cref{sec:outsideTraining}). For this evaluation, we used the MICE Slovene ELMo model described in \Cref{sec:outsideTraining}, as it outperformed all other models in previous tests. 
Figure \ref{fig:f1_scores} shows the distribution of $F_1$ scores across all the IEs in our SloIE dataset. The distribution shows that for the majority of IEs, MICE models achieve high $F_1$ scores above 0.8, while there are a few IEs with low recognition rate with $F_1 < 0.6$. In Table \ref{tab:expression_scores} we elaborate on these results and show the five best and worst recognizable IEs. At the moment, we do not have an interpretation of why certain IEs are more or less difficult to detect, and leave this question for further work.

\begin{table}[h!tb]
 \caption{Examples of the easiest and most difficult IEs and their direct/idiomatic translation for the MICE model with Slovene ELMo embeddings. }
  \centering
  \begin{tabular}{lcc}
    \toprule
    IE     & $F_1$ score & Number of detected IEs\\
    \midrule
    pospraviti v arhive & 1.0 & 4\\
    \multicolumn{3}{l}{\textit{to archive / to remove from attention }} \\
    kislo jabolko & 1.0 & 9\\
    \multicolumn{3}{l}{\textit{sour apple / unpleassant matter }} \\
    pomešati jabolka in hruške & 1.0 & 33\\
    \multicolumn{3}{l}{\textit{compare apples and pears / compare things that cannot be compared }} \\
    pristati v žepih nekoga & 1.0 & 28\\
    \multicolumn{3}{l}{\textit{lend in someones pockets / to steal }} \\
    perje začne frčati & 1.0 & 19\\
    \multicolumn{3}{l}{\textit{the feather starts to flutter / to make a public uproar }} \\
    
    \midrule
    pospraviti kaj v arhiv & 0.600 & 12\\
     \multicolumn{3}{l}{\textit{to archive something  / to end something }} \\
    imeti krompir & 0.597 & 162\\
     \multicolumn{3}{l}{\textit{to have a potato  / to be lucky }} \\
      gnilo jajce & 0.571 & 11\\
      \multicolumn{3}{l}{\textit{rotten egg / unpleasant surprise }} \\
    kdo nosi hlače & 0.525 & 218\\
    \multicolumn{3}{l}{\textit{to wear trousers  / to be in charge }} \\
    Želodec se obrne & 0.487 & 10\\
    \multicolumn{3}{l}{\textit{to turn the stomach  / to be disgusted}} \\
    \bottomrule
  \end{tabular}
  \label{tab:expression_scores}
\end{table}


\subsection{Cross-lingual evaluation of IEs}
\label{sec:XLevaluation}
The results above show encouraging results for IE detection in a language with sufficiently large datasets. As recent research on cross-lingual embeddings shows that reasonably good transfer of trained models can be obtained for many tasks \cite{ruder2019survey,artetxe2019massively,robnik-mozetic20,linhares2020}, we attempt such a transfer of our models. We use the dataset from the PARSEME shared task on automatic identification of verbal multiword expressions described in Section \ref{sec:parseme}. We evaluated two contextual embeddings discussed in the previous sections: the Slovene ELMo embeddings and the multilingual BERT embeddings. We evaluated the cross-lingual MICE approach in the following manner: 
\begin{itemize}
    \item We evaluated MICE with Slovene ELMo embeddings on Slavic languages similar to Slovene, with datasets present in the PARSEME collection, i.e., Slovene, Croatian, and Polish. As the Slovene ELMo embeddings are not multilingual, they are unlikely to generalize to other languages. In future work, we plan to use these embeddings for prediction in other languages by using cross-lingual mappings (e. g., \cite{schuster-etal-2019-cross}).
    \item We evaluated MICE with mBERT embeddings on all languages from the PARSEME collection. The mBERT model was trained on 104 languages, including every language present in the PARSEME dataset. 
\end{itemize}

For both test-cases, we constructed  balanced datasets which consist of every sentence with IEs from the PARSEME dataset in a given language, and an equal number of sentences without IEs, chosen at random from the same dataset. We performed the evaluation on the sentence-level classification task. 

For the Slavic languages test, we trained the prediction model on the whole SloIE dataset, presented in Section \ref{sec:monolingual-dataset}. We did not train the model on any multilingual data to see whether the contextual embeddings alone are enough to generalize to other languages, at least to similar ones such as Croatian. For all PARSEME languages using MICE with mBERT, we split each dataset into the training, testing and validation sets using a 60:30:10 ratio, trained the model for each language on the training set and evaluated it on the testing set. For Slovene, Croatian, and Polish we additionally trained MICE mBERT models on the SloIE dataset, as the similarity of those languages means that additional data in the Slovene language could be beneficial. The results are presented in Table \ref{tab:results_multilingual}.

\begin{table}[h!!tb]
 \caption{Results of the multilingual evaluation.  The MICE models with Slovene ELMo embeddings were evaluated on Slavic languages similar to Slovene, while the variants with mBERT were tested for all languages in PARSEME dataset which contain IEs. We report $F_1$ scores and include default classifiers as a reference.}
  \centering
  \begin{tabular}{lccc}
    \toprule
    Language & Slovene ELMo & mBERT  & Default $F_1$ \\
    \midrule
    Slovene & 0.8163  & 0.8359  & 0.667\\
    Croatian & 0.9191   & 0.8970 &  0.667\\
    Polish  & 0.2863 & 0.6987   &  0.667\\ 
    \midrule
    English  &  - & 0.650 & 0.667 \\ 
    French & - & 0.814 & 0.667 \\ 
    German & - & 0.622  & 0.667\\ 
    Turkish & - & 0.682 & 0.667\\ 
    Romanian & - & 0.625 & 0.667\\ 
    Lithuanian & - & 0.689 & 0.667\\ 
    Italian & - & 0.683 & 0.667\\ 
    Hungarian & - & 0.555 & 0.667\\ 
    Hindi & - & 0.562 & 0.667\\ 
    Hebrew & - & 0.693 & 0.667\\ 
    Farsi  & - & - & - \\
    Basque & - & 0.692 & 0.667\\ 
    Spanish & - & 0.340 & 0.667\\ 
    Greek & - &  0.484 & 0.667\\ 
    Bulgarian & - & 0.601 & 0.667\\ 
    \bottomrule
  \end{tabular}
  \label{tab:results_multilingual}
\end{table}

The results of the monolingual evaluation presented in \Cref{sec:outsideTraining} are also confirmed on the Slovene PARSEME dataset, as MICE with Slovene ELMO model is capable of detecting idioms in that dataset. The same model generalizes very well to the PARSEME Croatian dataset, likely due to its similarity to Slovene. The generalization to  Polish, which is more distant Slavic language, is not successful. MICE models with mBERT also generalize  well for a few languages. They obtain good results on Slovene and Croatian, likely due to the large amount of training data in the SloIE corpus, which also generalizes to Croatian idioms. The MICE mBERT models outperform default classifiers in French, Turkish, Lithuanian, Italian, Hebrew, and Basque, despite small amounts of training data, low numbers of IEs in training sets, most IEs only appearing once, and IEs in the testing set not appearing in the training set. They perform less well on other languages. even obtaining scores below the default classifier. This is expected, as the PARSEME dataset  only contains a small number of IEs, with only one or two sentences for each expression. This means that most of the idioms in the test set did not appear in the training set. Our evaluation on the much larger SloIE dataset shows that achieving good results on IEs outside the training set is difficult even when using a large training dataset. Therefore, the small size of the datasets for individual languages is the main reason that the models performed worse than the default classifier in several languages.  MUMULS and the SVM baseline were both unable to detect IEs in other languages, obtaining the $F_1$ score of 0 in all cases. \hl{We did not perform cross-lingual evaluation using ensemble models. The results of individual models show that there is only one case where an ensemble might be useful (Croatian with Slovene ELMo and mBERT), which has a small amount of data (3,003 sentences). In order to evaluate how ensemble models perform on cross-lingual classification, a more comprehensive analysis would be required, which we leave for future work.}

\subsection{Effect of the dataset size}
\label{sec:test_dataset_size}
Most languages currently do not have IE datasets, and it might be helpful to provide an information on how large datasets are required. In this section, we analyze the size of dataset needed to obtain acceptable performance in Slovene language and expect that findings will generalize to other languages. Further, as our SloIE dataset is larger than existing IE datasets, our results are not directly comparable to existing research, which was evaluated on smaller datasets. Our evaluation will shed light on this question as well.

We approach the analysis by running a number of tests on subsets of SloIE dataset. We randomly selected subsets of different sizes (100 \%, 80\%, 60\%, 40\%, 20\%, and 10\% percent) and re-ran the evaluations, repeating tests with IEs from the training set (\Cref{sec:inTraining}).
We only tested our best model, MICE, with Slovene ELMo embeddings. We show the results when classifying IEs from the training set in Table \ref{tab:dataset_size}. The results show that MICE performs well even when using smaller datasets. The $F_1$ score and CA slowly decrease with lower numbers of training sentences and remain quite high even with smaller training sets. This means that our approach could achieve good real-world performance even with languages that do not have large annotated datasets. When classifying IEs from outside the training set, the results did not significantly change with lower dataset sizes. 

\begin{table}[bht]
 \caption{The effect of dataset size on classification accuracy (CA) and $F_1$ score using the sentence-level classification task with IEs that appear in the training set, using MICE with Slovene ELMo embeddings.}
  \centering
  \begin{tabular}{lcc}
    \toprule
    Sentences    & CA & $F_1$ score\\
    \midrule
    27698 & 0.903 &  0.938\\ 
    17449 & 0.906 & 0.942\\ 
    9771 & 0.902 & 0.938\\ 
    4787 & 0.870 & 0.934\\ 
    2010 & 0.894  & 0.934\\ 
    703 & 0.874 & 0.924\\ 
    \bottomrule
  \end{tabular}
  \label{tab:dataset_size}
\end{table}

Our final evaluation checks whether a balanced dataset improves the result. The SloIE dataset is highly imbalanced (both in the number of examples per IE and in the number of idiomatic and literal use cases of each expression). This might make training neural networks difficult. To determine how much the dataset imbalance effects the results we constructed a smaller, balanced dataset, that contains the same amount of idiomatic/non-idiomatic sentences for each expression. The balanced version of the dataset contains 5481 training sentences and 2349 testing sentences across 75 IEs.  

The balanced dataset is much smaller than the original dataset, and possibly reduced performance may be due to a smaller amount training data.. For a more fair comparison, we also constructed a smaller, imbalanced dataset by taking a random subset of SloIE sentences for each expression equal in size to the balanced dataset. The size and number of sentences for the imbalanced dataset were the same as the balanced version.

We performed sentence-level classification on the two datasets, predicting IEs present in the training set. The results of the classification are shown in Table \ref{tab:balanced_dataset}. Results show that training the model on the balanced dataset did not lead to an improved classification accuracy or $F_1$ score. This indicates that MICE is insensitive to this sort of imbalance and performs well even when trained on imbalanced datasets.

\begin{table}[htb]
 \caption{The effect of using a balanced dataset on classification accuracy and $F_1$ score. The evaluation was conducted as a sentence-level classification task with IEs appearing in the training set, and using MICE with Slovene ELMo embeddings. }
  \centering
  \begin{tabular}{lcccc}
    \toprule
    Dataset & CA & $F_1$ score & Default CA & Default $F_1$\\
    \midrule
    Balanced & 0.8011 & 0.766 & 0.500 & 0.667\\
    Imbalanced & 0.812 & 0.853 & 0.625 & 0.767\\
    \bottomrule
  \end{tabular}
  \label{tab:balanced_dataset}
\end{table}


\section{Conclusion and Future Work}
\label{sec:conclusion}
We showed that contextual word embeddings can be used with neural networks to successfully detect IEs in text. When contextual embeddings (ELMo or mBERT) were used as the first layer of a neural network with the same architecture as the existing MUMULS approach, we were able to obtain much better results. While the existing approaches performed relatively well on the sentence-level classification of IEs that were present in the training set, they failed on token-level tasks and when detecting new IEs, not present in the training set. We showed that using fine-tuned contextual word embeddings allows the network to perform better on token-level classification and to successfully generalize to IEs that were not present in the training set. This opens an opportunity for the successful treatment of IEs in many downstream applications. We published our code and models under the CC licence\footnote{\url{https://github.com/TadejSkvorc/MICE}}.

We evaluated our MICE approach on the SloIE dataset, a new, large dataset of Slovene idioms, as well as on the existing multilingual PARSEME datasets. SloIE dataset, which we made publicly available\footnote{\url{http://hdl.handle.net/11356/1335}}, is larger than most of existing datasets, and should therefore be useful for further research into automatic idiom detection. Additionally, we evaluated how the size of the dataset affects the results and showed that our approaches perform well even when trained on smaller datasets. 

We show that contextual word embeddings are capable of generalizing to other languages. When dealing with similar language pairs (e. g., Slovene-Croatian), both the monolingual ELMo embeddings and the multilingual BERT embeddings were capable of detecting idioms in Croatian text when trained only on Slovene. The multilingual BERT model was able to detect idioms even in some more distant languages, though with reduced classification accuracy and $F_1$ scores. \hl{Finally, a Bayesian ensemble of our best models has further improved the results on sentence-level classification, indicating that the used contextual embeddings contain at least some complementary information about IEs.}


Our work could be improved and extended in multiple ways. We only used embeddings that were pretrained on general text and were not fine-tuned for the specific task of detecting idiomatic language. Several authors have shown \cite{li2019specializing, devlin2018bert} that specializing embeddings for specific tasks can improve results on a variety of NLP tasks. Several such approaches could be applied to our task and would likely further improve the performance. Additionally, we intentionally used a simple network architecture that could be improved in the future.  \hl{A further examination of Bayesian ensemble models is also required to determine why the MM model performs well on sentence-level classification but is unable to improve performance when used on token-level classification and when the embeddings in the test set are not present in the training set.}
Finally, to put our models into  a practical use, we intend to apply MICE models in the task of IE lexicon construction.

 \subsection*{Acknowledgements}
The research was supported  by the Slovene Research Agency through research core funding no. P6-0411 and P6-0215, as well as the projects J6-8256 and J6-2581.
This paper is supported by European Union’s Horizon 2020 Programme project EMBEDDIA (Cross-Lingual Embeddings for Less-Represented Languages in European News Media, grant no. 825153). 

The SloIE dataset was annotated by student annotators Kaja Žvanut, Tajda Liplin-Šerbetar, Karolina Zgaga and Tjaša Jelovšek. A part of it was also annotated by a non-native speaker Danijela Topić-Vizcaya.

\bibliography{references}  

\end{document}